# Probability as a Modal Operator[1]

Alan M. Frisch
Peter Haddawy[2]

University of Illinois
Dept of Computer Science
1304 W. Springfield
Urbana, IL 61801
Frisch@m.cs.uiuc.edu
Haddawy@m.cs.uiuc.edu

**Abstract**

This paper argues for a modal view of probability. The syntax and semantics of one particularly strong probability logic are discussed and some examples of the use of the logic are provided. We show that it is both natural and useful to think of probability as a modal operator. Contrary to popular belief in AI, a probability ranging between 0 and 1 represents a continuum between impossibility and necessity, not between simple falsity and truth. The present work provides a clear semantics for quantification into the scope of the probability operator and for higher-order probabilities. Probability logic is a language for expressing both probabilistic and logical concepts.

## 1 Introduction

Probability is typically treated as a metatheoretic concept. In other words, one talks in a metalanguage about the probabilities of object-language sentences. For example, this is the way Nilsson's Probabilistic Logic is constructed [Nilsson, 1986]. There are numerous advantages, however, to treating probability as a modal operator in the object language. By this we mean that probability is treated as a type of logical operator so that statements about probabilities may be combined freely with other statements in the object language. We will call any such language a *probability logic*. The formal language we will use will be first-order logic with the addition of the modal operator $P$, subscripted by a real number to denote the probability of a sentence or a pair of numbers to indicate the interval within which the probability lies.

Such a probability logic provides a language for explicitly reasoning about probability. It gives us the ability to write sentences with both logical and probabilistic components such as "Jane plays the saxophone and is probably Miss America."

$$\text{Plays-sax(Jane)} \wedge P_{.8}(\text{Miss-America(Jane)})$$

It allows us to formulate higher-order probability statements, e.g. "There is a 50% chance that the coin is biased 2/3 in favor of heads."

$$P_{.5}P_{.67}(Heads)$$

It enables us to write sentences that combine the above two characteristics, such as: "The probability that it is cloudy and that the chance of rain is 30% is .2."

$$P_{.2}(Cloudy \wedge P_{.3}(Rain))$$

---

[1] The authors would like to thank Steve Hanks, David Selig, Peter Cheeseman, and Carl Kadie for their helpful comments on earlier drafts of this paper. We would particularly like to thank Patrick Maher for taking the time to set us straight on a number of critical issues.

[2] This work was supported by the author's Cognitive Science/AI fellowship.

It gives us an unambiguous way of expressing the relationships among the scopes of quantifiers and probability operators as in: "There is an employee who is probably a thief."

$$\exists x Employee(x) \wedge P_{.7}(Thief(x))$$

As a last illustration, consider the following example inspired by a cartoon in the magazine *Discover* (June, 1985). A general explains two alternative SDI systems to the President by saying:

> Plan A stops each Soviet missile with an 80% chance.
> Plan B has an 80% chance of stopping all Soviet missiles.

These two statements make very different assertions about the plans they are describing. In particular, if the President is a rational man he will choose plan B over plan A. We would like a logic which can represent the meaning of both these statements, can make clear how they differ, and can show why plan B is preferable to plan A.

Probability is represented by a continuum of values between 0 and 1. It is a popular belief in AI that probability is closely related to first-order logic partially because probability values represent a continuum between truth and falsity.

> *In this paper we present a semantical generalization of ordinary first-order logic in which the truth values of sentences can range between 0 and 1.* - Nils Nilsson [1986]

> *Formally, probability can be regarded as a generalization of predicate calculus, where instead of the truth value of a formula given the evidence (context) having only the values 0 (false) or 1 (truth), it is generalized to a real number between 0 and 1.* - Peter Cheeseman [1985]

This view, however, is not quite correct. To shed light on what kind of continuum probability values represent, consider the three pairs of sentences:

1) The coin will either land heads or tails on the next toss.
2) The coin will land heads or the coin will land tails on the next toss.

3) The coin will necessarily land either heads or tails on the next toss.
4) The coin will necessarily land heads or the coin will necessarily land tails on the next toss.

5) The probability that the coin will land either heads or tails on the next toss is one.
6) The probability of landing heads on the next toss is one or the probability of landing tails on the next toss is one.

Clearly the first statement is true if and only if the second is true. The second pair of statements, however, are not equivalent. The third statement is certainly true but the fourth would imply that the coin is either two-headed or two-tailed. The question is: is the third pair of statements more closely related to the first or to the second pair? If it corresponds to the first pair then the fifth statement would have to imply the sixth statement. This is clearly not the case. It is, in fact, not the case for exactly the same reason that the third statement does not imply the fourth. Thus it is clear just from intuition that probability represents a continuum between necessity and impossibility not between simple truth and falsity. *Alethic logic*, the modal logic of necessity and possibility, captures exactly this distinction. It seems reasonable to expect a model theory for probability logic to be related in some way to a model theory for alethic logic.

We are not the first to suggest a modal view of probability. Nicholas Rescher [Rescher, 1962] was perhaps the first to examine the relationship between probability and modality. He formulated a logic in which a probability of one was interpreted as necessity and showed that in a finite possibility space, modal logic S5 is "appropriately regarded, nothing more than a propositional probability logic." Since then several philosophers, logicians, and mathematicians have discussed this interesting relationship [Danielsson, 1967; Gärdenfors, 1975; Halpern and McAllester, 1984; Gaifman, 1986; Ruspini, 1987]. The work most closely related to that presented in this paper is Fagin and Halpern's work on probability and belief [1988a; 1988b; 1988]. The present work contributes to the topic by presenting a more thorough discussion of the relationship between modal logic and probability and a model theory that can handle both quantification and higher-order probability.

The standard way to formalize the semantics of modal logics is in terms of Kripke structures [Kripke, 1963]. A Kripke structure is a model composed of a set of possible worlds and a binary accessibility relation between the worlds. A possible world can be considered to be a complete description of one possible reality or state of affairs. In an alethic logic, one world is accessible from a given world if it is considered possible with respect to that given world. Various alethic logics satisfying different properties may then be defined in terms of restrictions on the accessibility relations of the models. To define a semantics for probability logic we may construct our models, analogously to Kripke structures, in terms of probability distributions over possible worlds.

## 2 Coherence Constraints

There are a number of different properties we may wish a probability logic to satisfy. If we wish the logic to describe coherent beliefs, the logic should certainly satisfy the axioms of probability. This requirement is met by requiring the weights on the possible worlds to be probability distributions. Coherence may further require higher- and lower-order probabilities to satisfy certain mutual constraints. Two possible constraints suggested by Brian Skyrms [Skyrms, 1980, Appendix 2] are a minimal self-knowledge constraint (C1) and Miller's principle (MP), as listed below. Additionally, several researchers [H.E. Kyburg, 1987; Good, 1965] have suggested an expected value constraint (EV). Each subsequent constraint below entails the previous ones.

**C1:** $PR(pr(A) \in I) = 1 \rightarrow PR(A) \in I$, where $PR$ is the higher-order probability and $pr$ is the lower-order probability.

**EV:** The second-order probability should equal the expectation of the second-order probability applied to the first-order probability.

**MP:** $PR(A|pr(A) \in I) \in I$

Which constraints are applicable is a function of the interpretation we place on the probabilities. If we interpret higher and lower order probabilities as the rational degree of belief of an agent that does not know his own mind then C1 is a reasonable constraint. An agent that violates C1, is certain that his degree of belief in $A$ is $I$ although it isn't. Constraint MP, Miller's principle, is justified only if an agent's belief changes are the result of learning experiences.[3] A number of researchers

---

[3]Kyburg [1987] has recently dismissed the utility of higher-order probabilities by arguing that "higher-order probabilities can always be replaced by the marginal distributions of joint probability distributions." The argument claims that because EV is an unavoidable constraint, higher- and lower-order probabilities can be combined into a joint

111

have used diachronic Dutch book arguments in an attempt show MP to be a general requirement for dynamic rationality [van Fraassen, 1984; Goldstein, 1983]. These arguments essentially attempt to show that if you accept a certain system of bets concerning your present and future beliefs and you do not adhere to MP, you are open to a Dutch book. Levi, however, has recently shown these arguments to be invalid [Levi, 1986].

If $PR$ represents rational degree of belief and $pr$ represents objective probability then MP is a plausible rule for assimilating knowledge about objective probability. If $pr$ represents the probabilities of choosing a white ball from several different urns and $PR$ represents the probability of choosing each of the different urns (both objective probabilities) then MP is again plausible.

We would like to formulate a logic which satisfies the strongest of the constraints discussed above, i.e. Miller's principle. This constraint will be expressed in the form of restrictions on the models. Weaker logics will simply be obtainable by relaxing the restrictions. The problem we address then is how to construct the Kripke structures in such a way that all of the examples presented earlier can be modeled and such that the logic conforms to Miller's principle. We present a solution that satisfies these requirements in the form of a logic based on Gaifman's [1986] theory of higher-order probability and Nilsson's [1986] Probabilistic Logic. For purposes of explication we will only discuss the theory restricted to probabilities nested to depth two; however, the general theory is capable of dealing with probabilities nested to any depth. Using this model theory, we will discuss the relation between probability logic and alethic logic.

## 3 Syntax of $L_{mp}$

We now give the syntactic rules for the language $L_{mp}$. As mentioned in the introduction, the language is that of first-order logic with the addition of the modal operator $P$.[4] Some examples of syntactically correct sentences are

$$P_{.5}A$$

$$P_{.8}(P_{.2}A \wedge P_{.3}B)$$

Since we are only discussing sentences with probabilities nested to depth two, we can further simplify the presentation by explicitly indicating the nesting level. We will use $P2$ for the outer level probability operator and $P1$ for the inner probability operator. So $P1$ will always appear nested within a $P2$ operator. $P1$ refers to the first-order probability while $P2$ refers to the second-order probability. The above examples are then written as follows:

$$P2_{.5}A$$

$$P2_{.8}(P1_{.2}A \wedge P1_{.3}B)$$

## 4 Semantics of $L_{mp}$

The model theory presented here is based on Gaifman's [1986] theory of higher-order probability. We have retained as much of the terminology as possible but have modified parts in order to conform to standard logical semantics. We hope that the present formulation is somewhat more accessible to researchers in both uncertainty and logic.

---

probability space. The problem with this argument is that there are cases in which EV need not hold and furthermore representing the higher- and lower-order probabilities in the manner Kyburg suggests requires the stronger Miller's principle, not just EV.

[4]When talking about the knowledge of multiple agents, the P operator is superscripted by the name of an agent.

Our probability models have all the components of a Kripke structure except that the accessibility relation is replaced with probability distributions. Instead of having an accessibility relation for each modal operator, we have a distribution for each of the two operators, $P1$ and $P2$. Alethic logic gives us a set of worlds that are accessible from a given world; probability logic adds a probability distribution over that set of worlds. So in addition to saying that a world is possible, we say how possible.

A model is a five tuple $M = \langle W, PR2, PR1, D, F \rangle$, where $W$ is an arbitrary set of objects we will call the possible worlds, $PR2$ and $PR1$ are functions from the set of worlds $W$ to the set of probability distribution over $W$, $D$ is the domain of all individuals, and $F$ is the denotation *function* that takes a non-logical symbol and a world into the the denotation of that symbol in that world. $PR1$ and $PR2$ are the first- and second-order probability distributions, respectively. The first-order probability distribution associated with a world $w$ is denoted by $PR1^w$. Note that Nilsson's [1986] Probabilistic Logic models correspond to the special case in which there is only one distribution level and the probability distributions of all the worlds are identical. Furthermore, a world in Nilsson's models corresponds to an equivalence class of worlds in our models.

In the following discussion, we will have need to refer to the semantic value of a sentence before we have given the actual formal definitions for semantic value. The semantic value of a sentence $\alpha$ relative to a model $M$, a world $w$ in $M$, and an assignment $g$ of values to variables, will be denoted by $[\![\alpha]\!]^{M,w,g}$.

We now present the rules for determining the probability value of a sentence in a probabilistic model. The first-order *probability value* of a sentence $\alpha$ relative to a model, a world, and a value assignment is just the measure, with respect to the first-order probability distribution, of the set of worlds in which the sentence is true:

$$Prob1^{M,w,g}(\alpha) = PR1^w\{\hat{w} : [\![\alpha]\!]^{M,\hat{w},g} = T\}$$

As mentioned earlier, we would like our probability logic to satisfy Miller's principle. This will be done by first enforcing the expected value property and then adding a constraint to yield Miller's principle.[5] The expected value property could be achieved by placing a direct constraint on the second-order probability distribution:

$$PR2(\alpha) = \int_W PR2(w) \cdot PR1(\alpha)\, dw$$

and defining the second-order probability value in the same way as *Prob1* above. This is the way Gaifman [1986] constrains his models.

Alternatively, we have chosen to incorporate the expected value constraint into the rule for evaluating the second-order probability value. This gives us much more freedom in choosing the probability distributions for the models. The second-order probability value of a sentence will be defined as the expected value of the first-order probability value:

$$Prob2^{M,w,g}(\alpha) = \int_W PR1^{w'}\{\hat{w} : [\![\alpha]\!]^{M,\hat{w},g} = T\} \cdot PR2^w(w') dw'$$

There is one problem with the definitions of probability value given above: The set of worlds which satisfy a sentence may not be measurable. There are a number of possible solutions to this problem. If the number of possible worlds is either finite or countable then the set is always measurable and

$$Prob1^{M,w,g}(\alpha) = \sum_{\{w_i : [\![\alpha]\!]^{M,w_i,g} = T\}} PR1^w\{w_i\}.$$

---

[5]This then makes it clear how we could formulate a weaker probability logic.

| dist level | $w_1$ 2H'd,H | $w_2$ F,H | $w_3$ F,T | $w_4$ 2T'd,T |
|---|---|---|---|---|
| 2 | .2 .1 .2 .5 | .2 .3 0 .5 | .2 .2 .1 .5 | .2 0 .3 .5 |
| 1 | 1 0 0 0 | 0 .5 .5 0 | 0 .5 .5 0 | 0 0 0 1 |

Figure 1: Model for the coins example.

Nilsson [1986] solves the measurability problem by restricting his logic to finite sets of sentences with no logical operators or quantifiers outside the scope of the probability operator. In this way he needs only to define a measure over the finite number of equivalence classes determined by the maximum $2^n$ consistent truth assignments to the $n$ sentences. The probability of a sentence is then just the sum of the probabilities of the equivalence classes in which the sentence is true. Finally, Fagin and Halpern [1988b] discuss a number of possible solutions. For their propositional probability logic, they show that if every atomic formula is measurable then every formula is measurable. If a set is not measurable, they take the inner and outer measures of the set as roughly defining lower and upper bounds on the probability. They show that in the case of a finite probability space, the inner measure corresponds to a Dempster-Shafer belief function.

We need one additional constraint to enforce Miller's principle. Miller's principle is defined in terms of conditional probability. We will define the *conditional probability value* of a sentence $\alpha$ given $\beta$ in the usual way:

$$Prob2^{M,w,g}(\alpha|\beta) = \frac{Prob2^{M,w,g}(\alpha \wedge \beta)}{Prob2^{M,w,g}(\beta)}$$

When $Prob2^{M,w,g}(\beta) = 0$ the conditional probability value is undefined. The definition for $Prob1$ is similar.

Then Miller's principle says that

$$\frac{Prob2^{M,w,g}(\alpha \wedge P1_I(\alpha))}{Prob2^{M,w,g}(P1_I(\alpha))} \in I$$

The bottom term is just the expected value of the set of worlds in which the first-order probability value of $\alpha$ is in the interval $I$. Then for Miller's principle to hold, the numerator must be this value multiplied by some number in the interval $I$. This is achieved by the following *equivalence class constraint*.

> Let $C$ be the set of all worlds with a given $PR1$ distribution. Then for almost all $w_i \in C^6$ $PR1^{w_i}(C) = 1$.

The following simple example should help to clarify the meaning of the rather complex terminology just presented. Suppose you are given a bag of coins and are told that it contains 2 double-headed coins, 3 fair coins, and 5 double-tailed coins. We would like to know what the probability is that a coin drawn at random from the bag will land heads. There are many models consistent with this description. One possible probability model is shown in Figure 1. The figure shows the two distributions for each of four worlds. World $w_1$ represents the event that a two-headed coin is drawn and lands heads, world $w_2$ that a fair coin is drawn and lands heads, etc. Notice that worlds $w_1$ and $w_2$ have the same $PR1$ distribution so, consistent with the equivalence

---

[6] i.e. for all $w_1$ and $w_2$ for which there exists an $S \subset W$ and a $w$ such that $Prob2^{M,w,g}(S) = 1$ and $w_1, w_2 \in S$

class constraint, in each of them $PR1\{w_1, w_2\} = 1$. We calculate the second-order probability value of heads in world $w_1$ as the expected value of the worlds in which the coin lands heads:

$$Prob2^{w_1}(H) = .2(1) + .1(.5) + .2(.5) + .5(0) = .35$$

Since the probabilities in this example are objective probabilities, we would like Miller's principle to hold. We can see that this is the case:

$$Prob2^{M,w_1,g}(H \wedge P1._5H) = .15$$

$$Prob2^{M,w_1,g}(P1._5H) = .3$$

Thus from the definition of conditional probability value it follows that

$$Prob2^{M,w_1,g}(H|P1._5H) = .5$$

Using the above definitions, we can now define the semantic values of sentences of the language. As in alethic logics, the denotation of an expression is determined relative to a model, a world, and a value assignment of individuals from the domain to variables.

Let $M = \langle W, PR2, PR1, D, F \rangle$ be a model, $w$ a world, and $g$ a value assignment. Then $[\![\alpha]\!]^{M,w,g}$ is defined in the usual way for modal logic, with addition of the two rules:

- If $\alpha$ is $P2_I(\phi)$ then $[\![\alpha]\!]^{M,w,g} = T$ if $Prob2^{M,w,g}(\phi) \in I$.
  Otherwise $[\![\alpha]\!]^{M,w,g} = F$.

- If $\alpha$ is $P1_I(\phi)$ then $[\![\alpha]\!]^{M,w,g} = T$ if $Prob1^{M,w,g}(\phi) \in I$.
  Otherwise $[\![\alpha]\!]^{M,w,g} = F$.

A sentence is *satisfied* by a model and a world iff its semantic value in that model and world is $T$. A sentence is *valid* iff it is satisfied by every model and world.

To illustrate the use of the semantic definitions, we will now analyze the descriptions of the two SDI plans presented in the introduction. The two descriptions can be formally written as follows:

Plan A: $\forall x[Missile(x) \rightarrow P2._8Stopped(x)]$
Plan B: $P2._8(\forall x[Missile(x) \rightarrow Stopped(x)])$

We assume that *Missile* is a rigid designator, i.e. that the set of missiles is the same in each possible world. This is reasonable since we are not worried about not knowing whether or not an object is a missile. Now, why should the President prefer plan B? We will show that the worst case scenario for plan B is the best case for plan A so plan B is at least as good as plan A. By the semantic definitions, sentences A and B are true iff

(A) for all $d \in D[[\![Missile(x)]\!]^{M,w,g[d/x]} = F$ or $[\![P2._8Stopped(x)]\!]^{M,w,g[d/x]} = T]$

(B) $\int_W [PR1^{\hat{w}}\{\hat{w} :$ for each $d \in D[[\![Missile(x)]\!]^{M,\hat{w},g[d/x]} = F$ or
  $[\![Stopped(x)]\!]^{M,\hat{w},g[d/x]} = T]\} \cdot PR2^w(\hat{w})] = .8 \, d\hat{w}$

Since the set of missiles is the same in each world, we can eliminate $Missile(x)$ from the two semantic formulas and consider only models in which the domain contains only missiles. We can then expand these two formulas out to obtain:

115

(A) for all $d \in D[\int_W [PR1^{\acute{w}}\{\hat{w} : [\![Stopped(x)]\!]^{M,w,g[d/x]} = T\} \cdot PR2^w(\acute{w})] = .8\, d\acute{w}]$

(B) $\int_W [PR1^{\acute{w}}\{\hat{w} : \text{for each } d \in D[[\![Stopped(x)]\!]^{M,\hat{w},g[d/x]} = T]\} \cdot PR2^w(\acute{w})] = .8\, d\acute{w}$

So B holds in those models in which the probability of the worlds in which all missiles are stopped is .8. The remainder of the worlds assigned non-zero probability do not stop all of the missiles but may stop some of the missiles. So any given missile is stopped with probability at least .8. For plan A, each missile is stopped with probability exactly .8. Hence plan B is at least as good as plan A.

## 5 Relation To Modal Logic

We now formally examine the relationship between probability logic and modal logic. Note that we will use $\Box$ to denote necessity and $\Diamond$ to denote possibility. It was pointed out in the introduction that probabilistic certainty is closely related to necessity. It is, however, not identical with necessity. The reason is that in an uncountable probability space we can have possible events of measure zero.[7] For example, suppose I am going to pick a real number at random in the interval [0 1]. For each number, the probability that it will be picked is zero, yet it is possible that it will be picked. In other words, the first statement below is satisfiable, whereas the second statement is not.

$$\forall x P2_0 picked(x) \land P2_1 \exists x picked(x)$$

$$\forall x \Box \neg picked(x) \land \Box \exists x picked(x)$$

Necessity is a stronger notion than probability one. For a sentence to be necessary it must be true in all possible worlds. This is not the case for probability one. The remainder of this section concerns only finite probability spaces.

In finite models we can identify necessity with probability one. We create the *probabilistic translation* of an alethic sentence by replacing all necessity operators by probability one. Since any alethic sentence containing possiblity operators can be represented using only necessity operators, this is sufficient to handle all alethic sentences. The question then is to what alethic logic does our probability logic correspond. As we have seen, the various restrictions on probability values (the axioms of probability, Miller's principle) impose restrictions on the structure of probability models. So a given probability logic will not correspond to just any alethic logic. In particular, our probability models correspond to Kripke structures with accessibility relations which are both transitive and serial. The weakest alethic logic in which accessibility must be both transitive and serial is D4. For the general theory we have the following result:

> *Any alethic sentence is satisfiable by a finite Kripke structure of logic D4 iff its probabilistic translation is satisfiable by a finite probabilistic model.*

Thus, in the finite case, our probability logic is a generalization of modal logic D4. We will call this logic PD4 for probabilistic D4.

To illustrate the correspondence between D4 and PD4, we construct a mapping from probabilistic models to Kripke structures. Let $M = \langle W, PR2, PR1, D, F \rangle$ be a probabilistic model. This model can be mapped onto the Kripke structure $M' = \langle W', R, D, F \rangle$ as follows. For each world $w$

---

[7]Shimony [1955] has argued that it is irrational to assign probability zero to possible events. This has the intuitive appeal that an agent would always prefer to bet on a possible event over an impossible one. If we accept Shimony's argument then we can equate probability one with necessity. One possible way of modeling this is through the use of non-standard probabilities, based on non-standard analysis [Skyrms, 1980, Appendix 4].

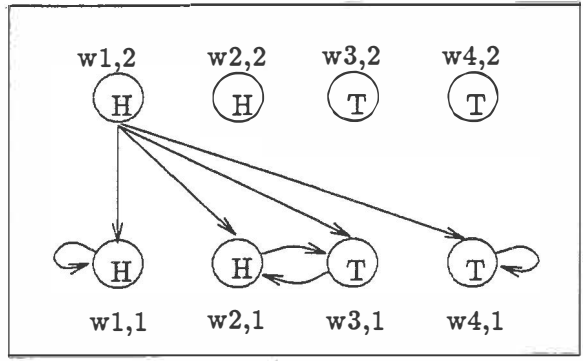

Figure 2: Kripke structure for coins example.

in $M$, we generate two possible worlds $\langle w, 1 \rangle$ and $\langle w, 2 \rangle$ in the Kripke structure, corresponding to $PR1$ and $PR2$ in the probabilistic model. Next we construct the accessibility relation $R$ according to the rule:

If $Prob2^{w_x}(\{w_y\}) > 0$ then $R(\langle w_x, 2 \rangle, \langle w_y, 1 \rangle)$ and
if $Prob1^{w_x}(\{w_y\}) > 0$ then $R(\langle w_x, 1 \rangle, \langle w_y, 1 \rangle)$.

The domain $D$ and denotation function $F$ of $M$ remain unchanged.

Figure 2 shows the Kripke structure for the probabilistic model of the coins example in Figure 1. Every world at level 2 is connected to every world at level 1. In world $\langle w_1, 1 \rangle$, $\Box H$ is true and in world $\langle w_4, 1 \rangle$ $\Box T$ is true. Thus in world $\langle w_1, 2 \rangle$, we have $\Diamond \Box H$, as well as $\Diamond \Box T$. This Kripke structure is a qualitative representation of the probabilistic model. Rather than representing degree of possibility now we can only represent possibility or impossibility.

## 6 Summary and Future Research

We have shown that, contrary to popular belief in AI, probability is related most closely to alethic logic, not first-order logic. A probability ranging between 0 and 1 represents a continuum between impossibility and necessity, not between simple falsity and truth.

We have discussed a number of possible constraints on higher-order probabilities. We presented a simplified version of a general probability logic which satisfies Miller's principle, the strongest of these constraints. This logic provides a clear semantics for higher-order probability and for quantification within and into the scope of the probability operator. Probability logic PD4 was shown to be a generalization of modal logic D4.

There are, of course modal logics both stronger than and weaker than D4. An interesting question is whether we can define probabilistic analogues of these other modal logics and whether these probability logics would have properties similar to their modal counterparts. In particular, knowledge logics such as S4 and S5 are characterized by the property that knowledge of a proposition implies the proposition is true. Probability is traditionally interpreted as representing belief. Could we define probability logics to represent knowledge?

A complete set of inference rules still needs to be developed for the logic. The relation of probability logic to alethic logic suggests the use of inference procedures similar to those used for modal logic. The clear correspondence of probability models to Kripke structures should help us understand how modal theorem proving methods might be applied or modified.

The theory thus far has only been developed for unconditional probabilities. In order to represent evidence and update beliefs, conditional probabilities are required. Conditional probability would be treated as a binary modal operator. There are difficulties in dealing with conditionals in a logical framework due the the fact that a conditional probability statement is undefined for conditioning statements with probability zero. One possible way of handling this is to make conditional probabilities primitive and define unconditional probabilities in terms of conditionals. This can be done using Popper functions [van Fraassen, 1976].